\documentclass{article}

\PassOptionsToPackage{numbers,sort&compress}{natbib}

 \usepackage[preprint]{neurips_2026}

\usepackage[utf8]{inputenc} 
\usepackage[T1]{fontenc}    
\usepackage[colorlinks=true,linkcolor=black,citecolor=black,urlcolor=black]{hyperref}
\usepackage{url}            
\usepackage{booktabs}       
\usepackage{amsfonts}       
\usepackage{nicefrac}       
\usepackage{microtype}      
\usepackage{xcolor}         
\usepackage{graphicx}
\usepackage{longtable}
\usepackage{flafter}
\usepackage{amsmath}
\usepackage{nicefrac}
\usepackage{microtype}
\usepackage{xcolor}
\usepackage{array}

\title{Robust to Which Model Change? A Unified Evaluation of Robust Counterfactual Explanations}
\workshoptitle{TAE (Trust-AI-Eval): Can We Trust AI Evaluation?}

\author{%
  Marcin Kostrzewa \\
  Department of Artificial Intelligence \\
  Wrocław University of \\ Science and Technology \\
  \texttt{marcin.kostrzewa@pwr.edu.pl}
  \And
  Maciej Zięba \\
  Department of Artificial Intelligence \\
  Wrocław University of \\ Science and Technology \\
  Tooploox \\
  \texttt{maciej.zieba@pwr.edu.pl}
}

\begin{document}

\maketitle

\begin{abstract}
Robust counterfactual explanations promise recourse that still works after the model behind it changes. Whether they keep that promise depends on what the change is. A small perturbation of the parameters, retraining on new data, and a new architecture are different events, and each existing method is evaluated against the one it was built for. Reported robustness scores, therefore, answer different questions and cannot be compared. We propose a unified cross-family evaluation protocol that holds factual instances and generated counterfactuals fixed while testing every method against the same eight types of model change. The benchmark compares six robust methods and two standard baselines on four tabular datasets. It characterizes every changed classifier through its outputs and reports empirical robustness together with coverage, base validity, and proximity. We find that relative performance and failure modes vary across change families. Bounded parameter perturbations change 0.95\% of test predictions on average, compared with 4.9\% for bootstrap retraining. Methods with guarantees for these perturbations do not necessarily transfer to other changes. RobX transfers most consistently in our experiments, although greater stability can require larger interventions. We argue that robust CFE methods should be evaluated through a common protocol that specifies the model changes, measures their realized behavioral magnitude, and keeps generation performance separate from robustness.
\end{abstract}

\section{Introduction}
\label{sec:introduction}

Counterfactual explanations (CFEs) are a common form of post-hoc explanation~\citep{Guidotti2022survey,Wachter2018method}.
Given an instance and a target outcome, a CFE describes feature changes under which the model
would produce that outcome. Depending on the application, CFEs may be assessed using desiderata
such as proximity to the original instance, plausibility, sparsity, and actionability~\citep{Guidotti2022survey,Verma2024survey}.
Most generation methods assess the CFEs for a fixed classifier.

This assumption is difficult to sustain when a recommendation takes time to implement. 
A lender may retrain its scoring model, incorporate new observations, or modify its training procedure before an applicant acts on the proposed changes.
The same CFE can then become invalid, even though the person followed the recommendation. 
Robust CFE methods seek explanations that remain valid after such a model change~\citep{Jiang2024survey}.

Robustness to model change is defined relative to a set or distribution of possible future classifiers.
Existing methods make different choices, including bounded parameter perturbations,
local stability proxies, distributional ambiguity sets, and ensembles of retrained models.
Their evaluations usually follow these method-specific choices. The literature therefore lacks a common protocol for testing whether different methods remain robust under the same model changes.

We propose a unified cross-family protocol and use it to study whether conclusions about CFE robustness transfer across model-change families.
The protocol covers four tabular datasets, five base neural networks per dataset, and
eight model-change families. The data split and factual instances remain
fixed, and all variants are evaluated. We characterize the magnitude of a change 
through prediction disagreement and probability differences, 
allowing comparisons between models with different architectures.

We compare six robust CFE methods and two standard baselines under the same models and factual instances. Empirical robustness is evaluated among base-valid CFEs and reported together with coverage, base validity, and proximity.

Our contributions are as follows:

\begin{itemize}
  \item We show that published robustness scores for CFE methods are not comparable, because each method is evaluated against changed models of its authors' choosing, and we propose a protocol that tests every method against the same held-out changed models, drawn from eight change families on four datasets.
  \item We make the evaluation auditable. Each changed model is described by how much it moves the base model's predictions, models used for tuning or calibration are kept apart from those used for evaluation, and coverage, base validity, robustness, and distance are reported separately so that a score cannot hide failed generations or large interventions.
  \item Applying the protocol to six robust methods and two baselines, we find that rankings change with the kind of model change, that guarantees for bounded parameter perturbations say little about retraining, and that a robustness score read without coverage and distance overstates two of the methods.
\end{itemize}

\section{Robustness to model change}
\label{sec:robustness}

\subsection{Problem setting}

Let $f_0 \colon \mathcal{X} \rightarrow [0,1]$ be a binary classifier and, for any classifier $f$, let $h_f(x)=\mathbb{I}[f(x)\geq 0.5]$ denote its predicted class. For a factual instance $x$ with $h_{f_0}(x)=0$, a counterfactual $x^{\mathrm{cf}}$ is valid if $h_{f_0}(x^{\mathrm{cf}})=1$. After the model changes, a new classifier $f$ replaces $f_0$. The counterfactual survives this change when $h_f(x^{\mathrm{cf}})=1$.

This definition still requires a description of which classifiers may replace $f_0$. For each change family $g$, we use a distribution $Q_g$ over changed classifiers. The robustness of a counterfactual $x^{\mathrm{cf}}$ to family $g$ is the probability that it remains valid,

\begin{equation}
  R_{Q_g}(x^{\mathrm{cf}}) = \Pr_{f\sim Q_g}\!\left[h_f(x^{\mathrm{cf}})=1\right].
  \label{eq:empirical-robustness}
\end{equation}

We report \emph{empirical robustness}, the mean of $R_{Q_g}$ over the base-valid CFEs a method produces. This is the validity-after-model-change measure commonly used in the literature. Coverage and base validity are reported separately, so a method cannot hide generation failures behind a robustness value.

Our common evaluation does not replace method-specific guarantees. When a method provides a formal certificate, we report it separately and interpret it only for the model changes covered by that certificate. To compare the realized changes across all families, we describe each changed model by its prediction disagreement and mean absolute probability difference from $f_0$. Unlike parameter distance, both quantities remain defined when the model architecture changes.

\subsection{Existing robustness assumptions}

The methods in our comparison encode future model changes in different ways, summarized in Table~\ref{tab:method-assumptions}. ROAR optimizes against bounded changes to the coefficients and intercept of a linear classifier~\citep{Upadhyay2021Roar}. For a nonlinear classifier, it applies the same objective to a local LIME surrogate~\citep{ribeiro2016lime}.

RBR samples feature vectors around a nearby point on the classifier's decision boundary and labels them using the current classifier~\citep{nguyen2022RBR}. It smooths the class-conditional samples into Gaussian mixtures and considers alternative mixtures within prescribed Wasserstein radii. For each candidate, it considers the allowed mixture that gives the largest adverse-to-favorable posterior odds, then minimizes this value.

RobX evaluates the current model on Gaussian perturbations around a candidate and scores local stability as the mean target-class probability minus its standard deviation~\citep{dutta2022robx}. Candidates below a selected threshold are moved toward a nearby favorable training point that passes the test. BetaRCE instead represents a user-defined space of admissible models through sampled classifiers and estimates a lower bound on the probability that a CFE remains valid over this space~\citep{stepka2025betarce}. Following the authors' procedure, we first obtain a CFE with growing spheres and then start a second growing-spheres search from that CFE to find a nearby candidate that satisfies the probabilistic robustness test.

RNCE and AP$\Delta$S reason directly about neural-network parameters. RNCE uses interval abstractions to certify a counterfactual against all parameter changes within a bounded set~\citep{jiang2024interval}. AP$\Delta$S samples bounded parameter perturbations and gives a probabilistic certificate with a selected confidence level~\citep{marzari2024APDeltaS}. Both guarantees concern models with the same architecture as the base network.

\begin{table}[htbp]
  \caption{Model-change assumptions and robustness formulations of the methods evaluated in this work, and the changed models used in their original evaluations.}
  \label{tab:method-assumptions}
  \centering
  \small
  \begin{tabular}{>{\raggedright\arraybackslash}p{0.12\linewidth}>{\raggedright\arraybackslash}p{0.31\linewidth}>{\raggedright\arraybackslash}p{0.19\linewidth}>{\raggedright\arraybackslash}p{0.25\linewidth}}
    \toprule
    Method & Representation of model change & Robustness formulation & Changed models in original evaluation \\
    \midrule
    ROAR-LIME & Bounded coefficients and intercept of a local linear surrogate & Objective under the most unfavorable allowed change & Retrained on temporally or geographically shifted data \\
    \addlinespace[3pt]
    RBR & Wasserstein ambiguity set around a Gaussian mixture & Objective under the most unfavorable allowed distribution & Retrained on shifted data \\
    \addlinespace[3pt]
    RobX & Local input stability used as a proxy for retraining & Stability threshold & Retrained after small changes to the training data \\
    \addlinespace[3pt]
    RNCE & Interval-bounded parameter changes & Validity certificate for every allowed change & Incremental, leave-one-out, and full retraining \\
    \addlinespace[3pt]
    AP$\Delta$S & Random bounded parameter perturbations & Probabilistic certificate & One model retrained on added data \\
    \addlinespace[3pt]
    BetaRCE & User-defined distribution of admissible classifiers & Probabilistic bound & Retrained with new data, hyperparameters, or architectures \\
    \bottomrule
  \end{tabular}
\end{table}

The nearest-neighbor and Wachter~\citep{Wachter2018method} baselines make no robustness assumption. They provide reference points for the extra distance incurred by robust methods and for the robustness that may arise without explicitly optimizing for it.

\subsection{Why a unified evaluation is needed}

Each of these methods was evaluated against changed models of its authors' choosing (last column of Table~\ref{tab:method-assumptions}). The evaluations also differ in how many changed models are used, often one per dataset, and in whether requests without a valid CFE enter the score. The reported numbers therefore describe different experiments and cannot be compared.

Our unified protocol instead holds the generated CFEs fixed and tests them against held-out classifiers from all eight change families. This design measures transfer beyond the changes assumed by the method while preserving the scope of its formal guarantees. Failure of RNCE or AP$\Delta$S under retraining, architecture changes, or data updates does not contradict certificates defined for bounded parameter perturbations. Conversely, robustness under bounded parameter perturbations does not imply robustness to other deployment updates. We therefore report each model-change family separately.

\section{Evaluation design}

\label{sec:evaluation}

Our unified protocol asks how the same counterfactual behaves under different model changes. For each dataset and each independently trained base model, a method generates one CFE for every selected factual.
We then hold that CFE fixed and evaluate it on every held-out variant in the corresponding change family.
To prevent evaluation leakage, none of these variants is used for CFE generation or hyperparameter selection. 
All experiments were run on a MacBook Pro with an Apple M4 Pro processor. Code and instructions for reproducing the benchmark are available at \url{https://github.com/genwro-ai/robust-counterfactual-methods-audit}.

\subsection{Datasets and base models}

We use four binary tabular datasets: Breast Cancer Wisconsin Diagnostic~\citep{Wolberg1993WDBC}, Pima Indians Diabetes~\citep{Smith1988Pima}, Wine Quality~\citep{Cortez2009Wine}, and HELOC~\citep{FICO2018HELOC}. Dataset-specific labels are mapped to a common convention in which 0 is the adverse outcome and 1 is the favorable outcome. We freeze one stratified split, shared by all five base models, into 50\% training data, a 10\% update pool, 10\% validation data, and 30\% test data. Features are standardized using statistics from the training split. The validation set controls early stopping and CFE hyperparameter selection. We draw up to 250 test instances predicted as adverse by each base model. The selected instances are shared by all methods for that model. Table~\ref{tab:appendix-datasets} summarizes dataset dimensions, realized split sizes, factual counts, and base-model test performance.

RNCE and AP$\Delta$S require neural-network base models, so we use the same neural architecture for every method. Each base classifier is a multilayer perceptron with two 32-unit ReLU layers and a sigmoid output. We use five independent initializations while keeping the data split and training protocol fixed.

\subsection{Model variants}

For every dataset and independently trained base model, we construct 25 variants in each of the eight families in Table~\ref{tab:model-variants}. This gives 200 changed models per base classifier and 1,000 per dataset. We evaluate all generated variants and report their balanced accuracy as descriptive information.

\begin{table}[t]
  \caption{Construction of the held-out model variants. Each row contains 25 variants per base classifier. All families except \emph{new initialization} reuse the base model's initialization seed, so they isolate the data or training change from initialization noise.}
  \label{tab:model-variants}
  \centering
  \small
  \begin{tabular}{p{0.22\linewidth}p{0.68\linewidth}}
    \toprule
    Family & Construction \\
    \midrule
    New initialization & Same data, architecture, and training configuration but different seed \\
    Bootstrap & Bootstrap resampling of the training set \\
    Data deletion & Random deletion of 1\%, 5\%, or 10\% of the training observations \\
    Data addition & Addition of 25\%, 50\%, 75\%, or 100\% of the reserved update pool \\
    Label update & Random label flips for 1\%, 5\%, or 10\% of the training observations \\
    Training configuration & Prespecified combinations of optimizer, learning rate, and weight decay \\
    Architecture & Prespecified alternatives in network width and depth \\
    Parameter perturbation & Random $\ell_\infty$ perturbations at five radii, with five directions per radius \\
    \bottomrule
  \end{tabular}
\end{table}

We compare every variant with its base model on the full test set. We compute their hard prediction disagreement and mean absolute difference in predicted probability. Hard disagreement records how often the predicted class changes, while probability MAE also captures shifts that do not cross the decision threshold. These output-based quantities apply to all eight families, including architecture changes for which aligned parameter distance is undefined. Their realized values are reported in Table~\ref{tab:appendix-change-magnitude}.

\subsection{Counterfactual methods}\label{sec:methods}

We evaluate the eight methods of Section~\ref{sec:robustness}. Below we describe the most important choices. The full configuration of every method is given in Appendix~\ref{app:method-config}.

No hyperparameter is chosen with access to a changed model. Wachter and ROAR-LIME are tuned over up to 25 adverse validation instances: we keep the configuration with the highest base-model validity, break ties by distance. The RobX stability threshold $\tau$ is dataset specific in the original work, so we set it per base model from the stability scores of favorable training instances and evaluate two configurations. The \emph{balanced} configuration uses their median and the \emph{robustness-first} configuration their 90th percentile.

AP$\Delta$S takes its parameter-shift radius as an input. We estimate it per base model as the largest $\ell_\infty$ parameter change over ten incremental updates, each a single mini-batch pass of Adam over a bootstrap sample of the reserved update pool starting from the base checkpoint, and run the authors' procedure with $\alpha=0.999$, $R=0.995$, and a 30-second limit per request.

Models used during generation never serve as evaluation models. BetaRCE generates with a separate ensemble of 32 bootstrap models that share the base initialization. Neither this ensemble nor the AP$\Delta$S calibration models appear among the held-out variants.

\subsection{Metrics and uncertainty}

\emph{Coverage} is the fraction of requested factuals for which a method returns a finite, changed candidate. \emph{Base validity} is the fraction of returned candidates that the base model classifies in the target class. Empirical robustness is the mean of $R_{Q_g}$ (Equation~\ref{eq:empirical-robustness}) over base-valid CFEs, with $Q_g$ uniform over the 25 held-out variants of family $g$. It is undefined when a method produces none. \emph{End-to-end robustness} is the product of coverage, base validity, and empirical robustness: the fraction of requests that yield a CFE that is valid for the base model and, on average over $Q_g$, remains valid for the changed model. Empirical robustness isolates the quality of the returned CFEs, whereas end-to-end robustness also charges a method for the requests on which it produced no valid CFE. For a base-valid counterfactual $x'$, proximity over the $d$ standardized features is measured as
\begin{equation}
    D(x,x') = \frac{1}{d}\sum_{j=1}^{d}\frac{\lvert x'_j-x_j\rvert}{s_j},
\end{equation}
where $s_j$ is the median absolute deviation of feature $j$ in the training data. We use its standard deviation when the median absolute deviation is zero.

We report each metric as the mean over the five independently trained base models, together with its standard deviation across them. The base model, not the individual CFE--variant evaluation, is the unit of replication, because every evaluation under one base model depends on that model.

\section{Results}
\label{sec:results}

\begin{figure}[t]
  \centering
  \includegraphics[width=0.88\linewidth]{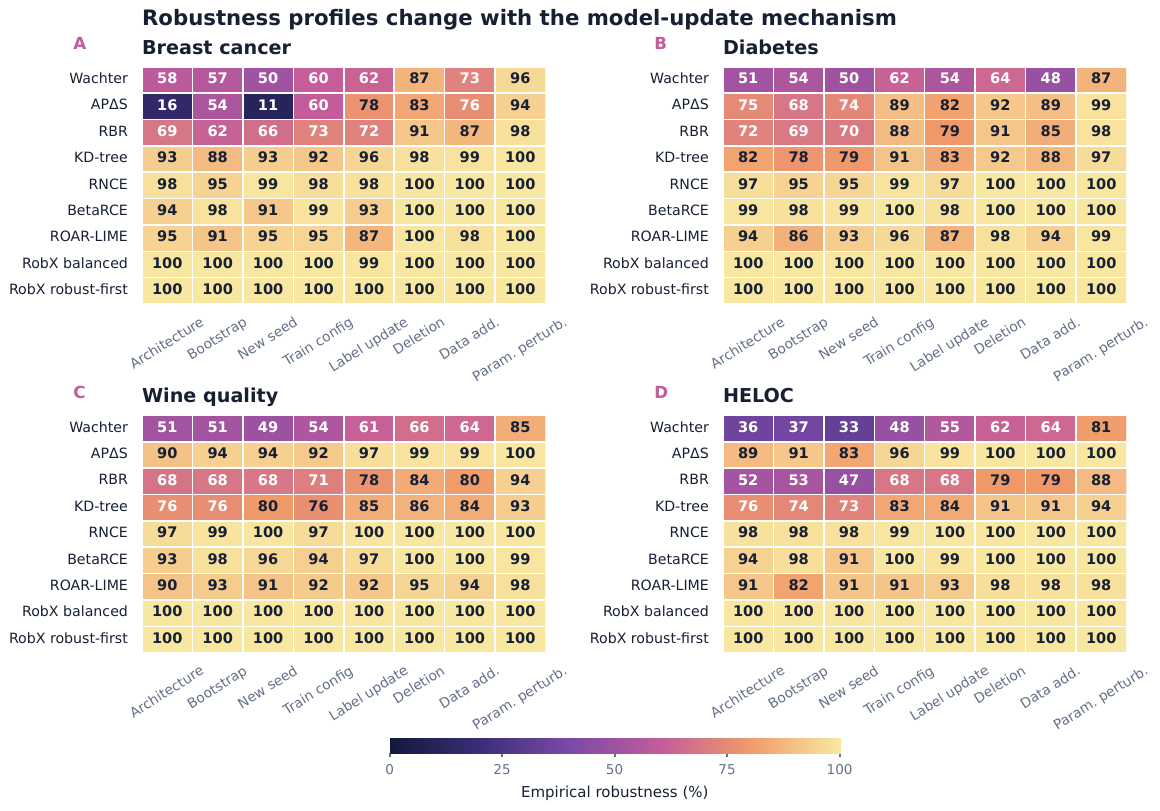}
  \caption{Empirical robustness across held-out model changes. Panels correspond to datasets, rows to CFE methods, and columns to the eight change families. Each cell aggregates 25 variants for each of five independently trained base models and measures changed-model validity among base-valid CFEs.}
  \label{fig:family-survival}
\end{figure}

Figure~\ref{fig:family-survival} shows empirical robustness by change family, Figure~\ref{fig:metric-decomposition} relates coverage, empirical robustness, and end-to-end robustness, and Figure~\ref{fig:proximity-tradeoff} plots empirical robustness against distance. Table~\ref{tab:appendix-full-results} in the appendix lists every value as mean $\pm$ standard deviation over the five base models.

\subsection{Robustness depends on the model-change family}

There is no stable ranking of methods across change families. The same CFE generator can perform well under a bounded parameter perturbation and fail after retraining. On Breast Cancer, AP$\Delta$S attains 94\% empirical robustness under bounded parameter perturbations, but only 11\% under a new initialization and 16\% under an architecture change. RBR ranges from 62\% under bootstrap retraining to 98\% under bounded parameter changes, and Wachter from 50\% under a new initialization to 96\% under bounded parameter changes.

The families also differ in how much they change the model (Table~\ref{tab:appendix-change-magnitude}). Bootstrap retraining alters 4.92\% of hard test predictions, architecture changes 4.69\%, and new initializations 4.57\%, compared with 0.95\% for bounded parameter perturbations. The uncertainty sets of the parameter-robust methods therefore cover the family that changes predictions least, which limits what their guarantees say about retraining or replacement.

Some methods are stable across the harder families. BetaRCE stays above 91\% in every dataset--family combination, RNCE never falls below 95\%, and the two RobX configurations never fall below 99.4\%.

\subsection{Robustness must be read with coverage and validity}

\begin{figure}[t]
  \centering
  \includegraphics[width=0.74\linewidth]{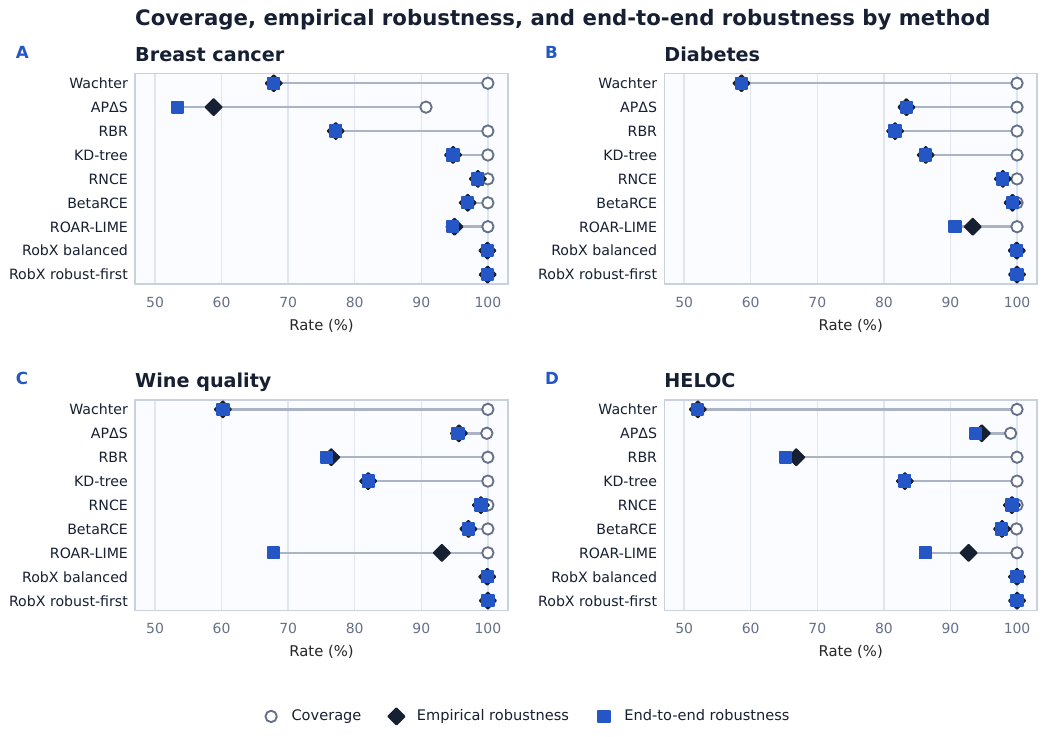}
  \caption{Coverage, empirical robustness, and end-to-end robustness for each method and dataset. The three markers coincide when a method returns a base-valid CFE for every request. The grey line spans the gap between them.}
  \label{fig:metric-decomposition}
\end{figure}

For most methods the three markers in Figure~\ref{fig:metric-decomposition} coincide: they return a base-valid CFE for every request, so empirical and end-to-end robustness are the same number. Two methods are different. ROAR-LIME returns a candidate for every factual, but on Wine Quality only 72.8\% of those candidates are valid for the base model, so its end-to-end robustness there is 67.8\% while its empirical robustness is 93.1\%. AP$\Delta$S returns no CFE for 9.6\% of Breast Cancer factuals, which lowers its end-to-end robustness on that dataset from 58.5\% to 53.1\%. All of these are 30-second timeouts of its margin search (43 requests in total, Table~\ref{tab:appendix-runtime}). The only other failure in the benchmark is one BetaRCE search on HELOC that exhausted its budget. A robustness score reported on its own would place ROAR-LIME near RNCE on Wine Quality. With coverage and base validity taken into account it lands below the nearest-neighbor baseline.

\subsection{Robustness has a price in distance}

\begin{figure}[tb]
  \centering
  \includegraphics[width=0.78\linewidth]{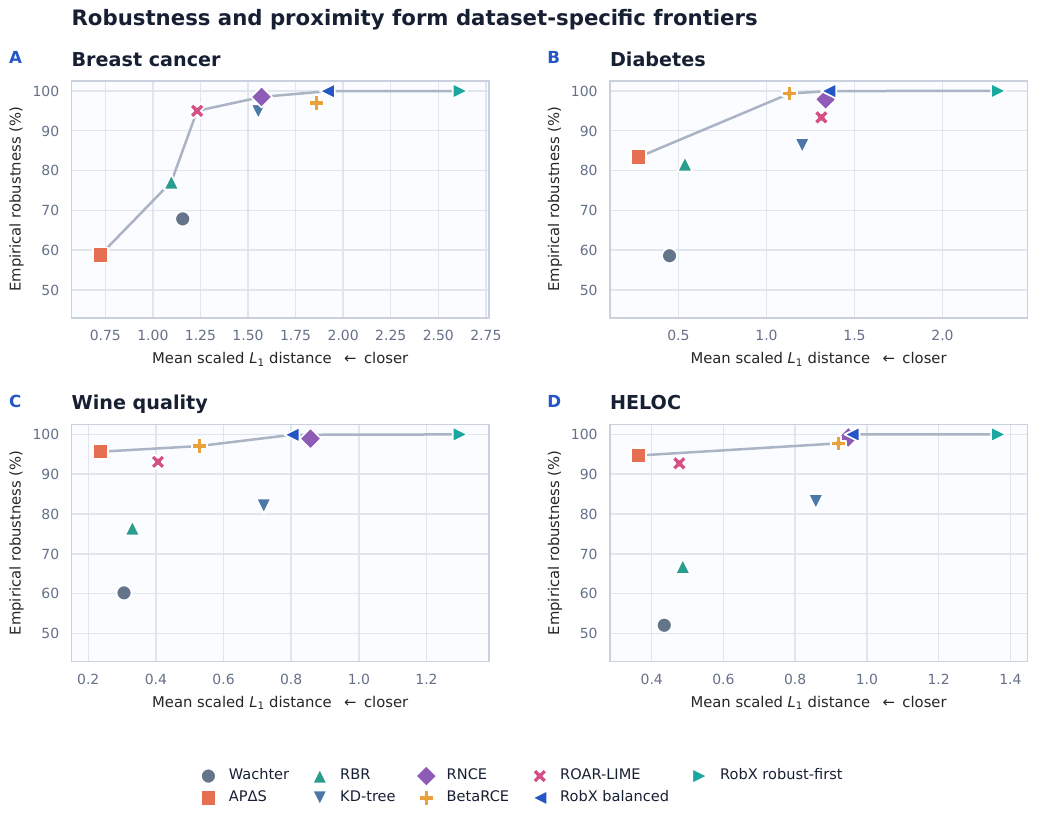}
  \caption{Proximity and empirical robustness by dataset. Each point is a five-seed aggregate, and pale lines trace the empirical proximity--robustness frontier. Both quantities are evaluated among base-valid CFEs.}
  \label{fig:proximity-tradeoff}
\end{figure}

On every dataset the closest method and the most robust method are different methods (Figure~\ref{fig:proximity-tradeoff}). AP$\Delta$S always produces the closest CFEs, and RobX robust-first always the most robust and the farthest ones. Between them, RobX balanced, RNCE, and BetaRCE form the high-robustness end of the frontier, with BetaRCE at the lowest distance of the three on Diabetes, Wine Quality, and HELOC. Distance alone does not buy robustness: the nearest-neighbor baseline moves as far as the robust methods and is nearly as robust as they are on Breast Cancer, but 12 to 18 points less robust on the other three.

The two RobX configurations show the price most clearly. Both anchor the counterfactual at a stable training instance and are equally robust, but the stricter threshold raises distance on every dataset, for example from 0.959 to 1.367 on HELOC.

\subsection{Formal guarantees retain their stated scope}

AP$\Delta$S is strongest under its own change model. Under bounded perturbations its empirical robustness is 94.1\% to 100.0\%, whereas over all families it is 58.5\% on Breast Cancer, 83.5\% on Diabetes, 95.7\% on Wine Quality, and 94.7\% on HELOC. Its independent post-selection check certifies 85.9\%, 98.9\%, 78.0\%, and 85.4\% of the requested CFEs, but the certificate refers to the calibrated parameter distribution and says nothing about retraining.

RNCE certifies its CFEs against parameter changes within a small interval. In this benchmark those CFEs also remain valid for 97.9\% to 99.2\% of the retrained and re-architected models. This is a property of our retraining procedures, not of the certificate. A different way of retraining could invalidate many of the same CFEs without contradicting the guarantee.

\section{Discussion}
\label{sec:discussion}

Which method to prefer depends on the update the deployed model is likely to receive and on how large an intervention is acceptable. Describing each change by how much it moves the model's predictions helps to read the results. Architecture changes, new initializations, bootstrap retraining, and label updates disagree with the base model most often, and they are the families on which methods built around small parameter neighborhoods lose the most. Two families with similar disagreement rates can still invalidate different counterfactuals, so output disagreement describes a change distribution but does not replace evaluating each family on its own.

\subsection{A unified evaluation protocol for robust counterfactuals}

We advocate a unified evaluation in which competing methods are tested against the same held-out model changes. A robustness claim should identify the changed-model set or sampling process before giving a score. The realized changes should then be described in output space, using at least hard disagreement and a continuous prediction difference. Parameter distance can be added when parameters are aligned, but it should not be used to compare architectures.

The CFE evaluation should keep distinct events separate: coverage, base validity, and empirical robustness answer different questions. End-to-end robustness combines the three rates into the fraction of requests that receive robust recourse and should be reported alongside them. Proximity must accompany these rates because robustness may be obtained through a larger intervention.

Finally, classifiers used to tune a method, construct an ensemble, or calibrate a certificate should be separate from the changed models used for evaluation. The experimental replicate is the independently trained base model, not every CFE--variant pair. Seed-wise means and standard deviations should reflect that hierarchy.

\subsection{Limitations}

The benchmark covers binary tabular tasks with numeric features and one adverse-to-favorable transition. It does not test multiclass outputs, images or text, causal feasibility, immutable features, or the time needed to carry out recourse. We evaluate proximity but do not evaluate other CFE desiderata such as sparsity, diversity, and plausibility.

We restrict the base classifiers to differentiable neural networks because RNCE and AP$\Delta$S require them. Including non-differentiable classifiers would reduce the set of methods that could be compared under the same models.

\section{Conclusion}

A counterfactual that is robust to one kind of model change need not be robust to another. In our benchmark, a method that survives small parameter perturbations can lose most of its counterfactuals once the model is simply retrained, and the methods that hold up under retraining usually ask the person for a larger change. A robustness claim is therefore only as informative as the description of the model change it was tested against. We propose to evaluate competing methods against the same held-out changed models, to report the result separately for each kind of change, and to keep coverage, base validity, robustness, and distance apart so that none of them hides the others. Results reported in this way can be compared across methods and papers, and they tell a practitioner what to expect from the update their own model will receive.

\section*{Acknowledgments}

This study was supported by the National Science Centre (Poland) Grant No. 2024/55/B/ST6/02100.

\bibliographystyle{unsrtnat}
\bibliography{bib}


\appendix

\section{Additional experimental details}

\subsection{Datasets, splits, and base classifiers}

All five base-model runs use the same fixed data partition. Because the evaluation considers the adverse-to-favorable transition, eligible factuals are test instances that the corresponding base model predicts as adverse. We use every eligible factual for Breast Cancer and Diabetes. For the larger Wine Quality and HELOC test sets, we draw a reproducible sample of 250 eligible factuals per base model. Table~\ref{tab:appendix-datasets} summarizes the datasets, splits, factual counts, and base-model performance.

\begin{table}[h]
  \caption{Dataset and base-model summary. $n$ and $p$ denote the numbers of observations and input features. Split sizes are reported as training/update/validation/test. ``Factuals'' gives the total number of adverse test predictions evaluated across the five base models, followed by the per-model range in parentheses. Test balanced accuracy is reported as minimum/mean/maximum across the five models.}
  \label{tab:appendix-datasets}
  \centering
  \small
  \begin{tabular}{lrrlrl}
    \toprule
    Dataset & $n$ & $p$ & Split sizes & Factuals & Test balanced accuracy \\
    \midrule
    Breast Cancer & 569 & 30 & 284/57/57/171 & 312 (60--64) & 0.967/0.973/0.988 \\
    Diabetes & 768 & 8 & 384/76/77/231 & 281 (50--61) & 0.712/0.721/0.736 \\
    Wine Quality & 6,497 & 11 & 3248/649/650/1950 & 1,250 (250) & 0.735/0.741/0.749 \\
    HELOC & 8,291 & 20 & 4145/829/829/2488 & 1,250 (250) & 0.727/0.730/0.735 \\
    \bottomrule
  \end{tabular}
\end{table}

Standardization parameters are fitted on the training partition. Every base classifier has two hidden layers of 32 ReLU units and a sigmoid output. We train the networks with full-batch Adam, a learning rate of $10^{-3}$, and no weight decay for at most 300 epochs. Early stopping monitors validation binary cross-entropy with patience 30 and restores the best epoch. The five training runs differ in model initialization while sharing the data partition and training protocol. For a given base model, every CFE method receives the same selected factuals.

\subsection{Realized magnitude of model changes}

Table~\ref{tab:appendix-change-magnitude} describes the 25 models in each family through their outputs on the test set. Each model appears only once in this calculation, although it is subsequently evaluated against every CFE. Bounded parameter perturbations use radii $\{0.001, 0.005, 0.01, 0.02, 0.05\}$ with five random directions each. A direction is drawn uniformly and scaled so that its largest component equals the radius, so realized perturbations are typical rather than adversarial members of the box. The training-configuration family combines Adam with learning rates $\{2.5\cdot10^{-4}, 5\cdot10^{-4}, 10^{-3}, 2\cdot10^{-3}, 4\cdot10^{-3}\}$ and weight decay $\{0, 10^{-5}, 10^{-4}, 10^{-3}\}$, excluding the base setting (19 variants), with SGD with momentum 0.9, learning rates $\{0.01, 0.03\}$, and weight decay $\{0, 10^{-4}, 10^{-3}\}$ (6 variants).

\begin{table}[h]
  \caption{Output difference from the base classifier: mean $\pm$ standard deviation over the 125 variants per dataset, macro-averaged across datasets.}
  \label{tab:appendix-change-magnitude}
  \centering
  \small
  \begin{tabular}{lrr}
    \toprule
    Change family & Hard disagreement (\%) & Probability MAE \\
    \midrule
    Bootstrap & 4.92 $\pm$ 1.25 & 0.039 $\pm$ 0.010 \\
    Architecture & 4.69 $\pm$ 1.54 & 0.038 $\pm$ 0.017 \\
    New initialization & 4.57 $\pm$ 0.93 & 0.036 $\pm$ 0.005 \\
    Label update & 3.69 $\pm$ 1.99 & 0.045 $\pm$ 0.025 \\
    Training configuration & 3.10 $\pm$ 1.99 & 0.028 $\pm$ 0.019 \\
    Data addition & 2.18 $\pm$ 0.96 & 0.017 $\pm$ 0.007 \\
    Data deletion & 1.87 $\pm$ 1.10 & 0.014 $\pm$ 0.008 \\
    Bounded parameter perturbation & 0.95 $\pm$ 1.17 & 0.007 $\pm$ 0.008 \\
    \bottomrule
  \end{tabular}
\end{table}

\subsection{Method configuration and separation of model pools}\label{app:method-config}

Wachter is tuned over $\lambda\in\{0.01,0.1,1\}$ and learning rates $\{0.01,0.02\}$. ROAR-LIME uses a perturbation radius of 0.1, 20,000 LIME samples, learning rate 0.001, and a validation search over its objective weight. Both searches maximize base-model validity and break ties by lower robust-scale $\ell_1$ distance. The validity-first ordering follows the ROAR authors' selection principle. RBR uses the midpoints of the parameter ranges considered by its authors: 1,000 candidate counterfactuals, 200 samples, a sampling radius equal to 0.2 of the maximum pairwise training distance (the authors' convention), and 500 optimization iterations. RNCE uses parameter and bias radii of 0.005. RobX uses variance 0.01 and 1,000 stability samples. For each base model, its balanced and robustness-first thresholds are the median and 90th percentile of stability among target-class training instances. Across seeds, the two thresholds range from 0.989 to 0.997 and 0.999 to 1.000 on Breast Cancer, 0.783 to 0.834 and 0.912 to 0.973 on Diabetes, 0.799 to 0.813 and 0.958 to 0.968 on Wine Quality, and 0.741 to 0.746 and 0.881 to 0.891 on HELOC. 

Following the BetaRCE paper, we use 32 bootstrap models, robustness threshold $\delta=0.9$, and confidence $\alpha=0.95$ for the one-sided lower credible bound. The bootstrap samples differ while initialization is fixed to that of the base model. Both growing-spheres stages use 100 samples per iteration, $\ell_2$ distance, and step size 0.1. We allow 200 iterations for the preliminary CFE search to avoid search-budget failures, while the second robustness search retains the authors' limit of 100 iterations.

For AP$\Delta$S, each of ten update replicas starts from a copy of the base checkpoint and receives a single pass of mini-batch Adam (learning rate $10^{-3}$, batch size 8, fresh optimizer state) over a bootstrap sample of the reserved update pool, drawn with replacement and of the same size as the pool. The radius is the maximum parameter $\ell_\infty$ distance among these replicas. Across the five seeds, selected radii range from 0.007494 to 0.007743 on Breast Cancer, 0.009563 to 0.009802 on Diabetes, 0.034274 to 0.050010 on Wine Quality, and 0.046518 to 0.049291 on HELOC. Generation follows the authors' iterative margin search with confidence 0.999 and target robustness probability 0.995, which require 1,378 sampled perturbations per robustness test, a 30-second wall-time limit per request, and an independent post-selection certification sample. Calibration replicas, the BetaRCE generation ensemble, and the 200 held-out evaluation variants have no models in common.

As a diagnostic, we also estimated the AP$\Delta$S radius from ten independently initialized BetaRCE bootstrap models. The resulting radii were 0.621--0.706 on Breast Cancer and 0.890--0.956 on Diabetes. AP$\Delta$S returned no CFE for any of five factuals under each of three seeds on either dataset. This diagnostic is deliberately small and is not part of the main comparison. It shows that a parameter radius is meaningful only relative to the process that produced it: independently initialized models can be close in predictions while far apart in raw parameter space.

\subsection{Generation time}

Table~\ref{tab:appendix-runtime} reports observed wall time for one CFE-generation request. The statistics pool all 3,093 requests per method across datasets and include unsuccessful attempts. They exclude work completed before the timed generation call, including model training, AP$\Delta$S calibration, and BetaRCE ensemble construction. Lazy initialization performed inside a generation call is included. In particular, RNCE constructs and caches its robust candidate set during the first request for each base model, which explains the difference between its mean and median. These measurements describe our implementations on the hardware reported in Section~\ref{sec:evaluation}, rather than implementation-independent computational complexity.

\begin{table}[h]
  \caption{Per-factual CFE generation time in seconds, reported as mean $\pm$ standard deviation across requests. AP$\Delta$S timeouts are unsuccessful requests that reach its 30-second limit.}
  \label{tab:appendix-runtime}
  \centering
  \small
  \begin{tabular}{lrrrr}
    \toprule
    Method & Mean $\pm$ SD & Median & 90th percentile & Timeouts \\
    \midrule
    AP$\Delta$S & 3.909 $\pm$ 4.643 & 2.612 & 7.542 & 43 \\
    BetaRCE & 0.0198 $\pm$ 0.0134 & 0.0159 & 0.0359 & 0 \\
    KD-tree & 0.0037 $\pm$ 0.0022 & 0.0042 & 0.0050 & 0 \\
    RBR & 1.754 $\pm$ 0.909 & 1.860 & 2.900 & 0 \\
    RNCE & 0.262 $\pm$ 4.200 & 0.0007 & 0.0008 & 0 \\
    ROAR-LIME & 0.086 $\pm$ 0.060 & 0.074 & 0.151 & 0 \\
    RobX balanced & 0.040 $\pm$ 0.028 & 0.030 & 0.065 & 0 \\
    RobX robust-first & 0.078 $\pm$ 0.034 & 0.073 & 0.123 & 0 \\
    Wachter & 0.0058 $\pm$ 0.0371 & 0.0036 & 0.0108 & 0 \\
    \bottomrule
  \end{tabular}
\end{table}

\section{Full dataset-wise results}

Table~\ref{tab:appendix-full-results} reports every dataset--method result. Rates are percentages. End-to-end robustness is the product of coverage, base validity, and empirical robustness. Every entry is the mean and standard deviation over the five independently trained base models. Distance is the mean robust-scale-normalized $\ell_1$ change among base-valid CFEs.

{\footnotesize
\setlength\tabcolsep{3.8pt}
\begin{longtable}{lrrrrr}
  \caption{Full five-seed results, reported as mean $\pm$ standard deviation over the five base models.}\label{tab:appendix-full-results}\\
  \toprule
  Method & Coverage $\uparrow$ & Base valid $\uparrow$ & Emp. robust. $\uparrow$ & E2E robust. $\uparrow$ & Distance $\downarrow$ \\
  \midrule
  \endfirsthead
  \multicolumn{6}{c}{\tablename\ \thetable\ continued}\\
  \toprule
  Method & Coverage $\uparrow$ & Base valid $\uparrow$ & Emp. robust. $\uparrow$ & E2E robust. $\uparrow$ & Distance $\downarrow$ \\
  \midrule
  \endhead
  \midrule
  \multicolumn{6}{r}{Continued on next page}\\
  \endfoot
  \bottomrule
  \endlastfoot
  \multicolumn{6}{l}{\textit{Breast Cancer}} \\
  AP$\Delta$S & 90.4 $\pm$ 19.8 & 100.0 $\pm$ 0.0 & 58.5 $\pm$ 3.6 & 53.1 $\pm$ 12.8 & 0.689 $\pm$ 0.201 \\
  BetaRCE & 100.0 $\pm$ 0.0 & 100.0 $\pm$ 0.0 & 97.0 $\pm$ 0.8 & 97.0 $\pm$ 0.8 & 1.860 $\pm$ 0.052 \\
  KD-tree & 100.0 $\pm$ 0.0 & 100.0 $\pm$ 0.0 & 94.8 $\pm$ 2.4 & 94.8 $\pm$ 2.4 & 1.554 $\pm$ 0.006 \\
  RBR & 100.0 $\pm$ 0.0 & 100.0 $\pm$ 0.0 & 77.1 $\pm$ 2.4 & 77.1 $\pm$ 2.4 & 1.097 $\pm$ 0.022 \\
  RNCE & 100.0 $\pm$ 0.0 & 100.0 $\pm$ 0.0 & 98.5 $\pm$ 0.8 & 98.5 $\pm$ 0.8 & 1.571 $\pm$ 0.010 \\
  ROAR-LIME & 100.0 $\pm$ 0.0 & 99.7 $\pm$ 0.7 & 94.9 $\pm$ 3.2 & 94.6 $\pm$ 3.8 & 1.232 $\pm$ 0.011 \\
  RobX balanced & 100.0 $\pm$ 0.0 & 100.0 $\pm$ 0.0 & 99.9 $\pm$ 0.1 & 99.9 $\pm$ 0.1 & 1.917 $\pm$ 0.056 \\
  RobX robust-first & 100.0 $\pm$ 0.0 & 100.0 $\pm$ 0.0 & 100.0 $\pm$ 0.0 & 100.0 $\pm$ 0.0 & 2.616 $\pm$ 0.051 \\
  Wachter & 100.0 $\pm$ 0.0 & 100.0 $\pm$ 0.0 & 67.8 $\pm$ 3.7 & 67.8 $\pm$ 3.7 & 1.157 $\pm$ 0.023 \\
  \addlinespace
  \multicolumn{6}{l}{\textit{Diabetes}} \\
  AP$\Delta$S & 100.0 $\pm$ 0.0 & 100.0 $\pm$ 0.0 & 83.5 $\pm$ 6.4 & 83.5 $\pm$ 6.4 & 0.276 $\pm$ 0.028 \\
  BetaRCE & 100.0 $\pm$ 0.0 & 100.0 $\pm$ 0.0 & 99.3 $\pm$ 0.3 & 99.3 $\pm$ 0.3 & 1.132 $\pm$ 0.053 \\
  KD-tree & 100.0 $\pm$ 0.0 & 100.0 $\pm$ 0.0 & 85.9 $\pm$ 5.9 & 85.9 $\pm$ 5.9 & 1.201 $\pm$ 0.058 \\
  RBR & 100.0 $\pm$ 0.0 & 100.0 $\pm$ 0.0 & 81.7 $\pm$ 5.7 & 81.7 $\pm$ 5.7 & 0.538 $\pm$ 0.045 \\
  RNCE & 100.0 $\pm$ 0.0 & 100.0 $\pm$ 0.0 & 97.9 $\pm$ 0.7 & 97.9 $\pm$ 0.7 & 1.336 $\pm$ 0.078 \\
  ROAR-LIME & 100.0 $\pm$ 0.0 & 97.3 $\pm$ 5.0 & 93.2 $\pm$ 2.5 & 90.8 $\pm$ 5.5 & 1.314 $\pm$ 0.069 \\
  RobX balanced & 100.0 $\pm$ 0.0 & 100.0 $\pm$ 0.0 & 100.0 $\pm$ 0.1 & 100.0 $\pm$ 0.1 & 1.358 $\pm$ 0.104 \\
  RobX robust-first & 100.0 $\pm$ 0.0 & 100.0 $\pm$ 0.0 & 100.0 $\pm$ 0.0 & 100.0 $\pm$ 0.0 & 2.316 $\pm$ 0.177 \\
  Wachter & 100.0 $\pm$ 0.0 & 100.0 $\pm$ 0.0 & 58.6 $\pm$ 2.6 & 58.6 $\pm$ 2.6 & 0.447 $\pm$ 0.049 \\
  \addlinespace
  \multicolumn{6}{l}{\textit{Wine Quality}} \\
  AP$\Delta$S & 99.8 $\pm$ 0.4 & 100.0 $\pm$ 0.0 & 95.7 $\pm$ 0.9 & 95.5 $\pm$ 0.9 & 0.236 $\pm$ 0.022 \\
  BetaRCE & 100.0 $\pm$ 0.0 & 100.0 $\pm$ 0.0 & 97.1 $\pm$ 0.4 & 97.1 $\pm$ 0.4 & 0.528 $\pm$ 0.011 \\
  KD-tree & 100.0 $\pm$ 0.0 & 100.0 $\pm$ 0.0 & 82.0 $\pm$ 4.1 & 82.0 $\pm$ 4.1 & 0.719 $\pm$ 0.021 \\
  RBR & 100.0 $\pm$ 0.0 & 99.0 $\pm$ 0.5 & 76.5 $\pm$ 3.8 & 75.7 $\pm$ 3.6 & 0.331 $\pm$ 0.014 \\
  RNCE & 100.0 $\pm$ 0.0 & 100.0 $\pm$ 0.0 & 98.9 $\pm$ 0.7 & 98.9 $\pm$ 0.7 & 0.857 $\pm$ 0.019 \\
  ROAR-LIME & 100.0 $\pm$ 0.0 & 72.8 $\pm$ 4.9 & 93.1 $\pm$ 0.9 & 67.8 $\pm$ 4.5 & 0.407 $\pm$ 0.012 \\
  RobX balanced & 100.0 $\pm$ 0.0 & 100.0 $\pm$ 0.0 & 99.9 $\pm$ 0.1 & 99.9 $\pm$ 0.1 & 0.802 $\pm$ 0.044 \\
  RobX robust-first & 100.0 $\pm$ 0.0 & 100.0 $\pm$ 0.0 & 100.0 $\pm$ 0.0 & 100.0 $\pm$ 0.0 & 1.300 $\pm$ 0.046 \\
  Wachter & 100.0 $\pm$ 0.0 & 100.0 $\pm$ 0.0 & 60.2 $\pm$ 5.9 & 60.2 $\pm$ 5.9 & 0.306 $\pm$ 0.016 \\
  \addlinespace
  \multicolumn{6}{l}{\textit{HELOC}} \\
  AP$\Delta$S & 99.0 $\pm$ 1.9 & 100.0 $\pm$ 0.0 & 94.7 $\pm$ 1.0 & 93.8 $\pm$ 2.6 & 0.363 $\pm$ 0.060 \\
  BetaRCE & 99.9 $\pm$ 0.2 & 100.0 $\pm$ 0.0 & 97.8 $\pm$ 0.5 & 97.7 $\pm$ 0.3 & 0.920 $\pm$ 0.031 \\
  KD-tree & 100.0 $\pm$ 0.0 & 100.0 $\pm$ 0.0 & 83.1 $\pm$ 1.2 & 83.1 $\pm$ 1.2 & 0.858 $\pm$ 0.018 \\
  RBR & 100.0 $\pm$ 0.0 & 97.6 $\pm$ 0.7 & 66.8 $\pm$ 2.3 & 65.2 $\pm$ 2.4 & 0.487 $\pm$ 0.021 \\
  RNCE & 100.0 $\pm$ 0.0 & 100.0 $\pm$ 0.0 & 99.2 $\pm$ 0.4 & 99.2 $\pm$ 0.4 & 0.948 $\pm$ 0.029 \\
  ROAR-LIME & 100.0 $\pm$ 0.0 & 93.0 $\pm$ 2.1 & 92.7 $\pm$ 0.7 & 86.2 $\pm$ 1.8 & 0.477 $\pm$ 0.030 \\
  RobX balanced & 100.0 $\pm$ 0.0 & 100.0 $\pm$ 0.0 & 100.0 $\pm$ 0.0 & 100.0 $\pm$ 0.0 & 0.959 $\pm$ 0.033 \\
  RobX robust-first & 100.0 $\pm$ 0.0 & 100.0 $\pm$ 0.0 & 100.0 $\pm$ 0.0 & 100.0 $\pm$ 0.0 & 1.367 $\pm$ 0.030 \\
  Wachter & 100.0 $\pm$ 0.0 & 99.9 $\pm$ 0.2 & 52.1 $\pm$ 3.0 & 52.0 $\pm$ 2.9 & 0.435 $\pm$ 0.021 \\
\end{longtable}
}



\end{document}